\documentclass[10pt,twocolumn,letterpaper]{article}

\usepackage{iccv}
\usepackage{times}
\usepackage{epsfig}
\usepackage{graphicx}
\usepackage{amsmath}
\usepackage{amssymb}

\usepackage{multirow}
\usepackage{booktabs}
\usepackage[breaklinks=true,bookmarks=false]{hyperref}

\iccvfinalcopy % *** Uncomment this line for the final submission

\def\iccvPaperID{6385} % *** Enter the ICCV Paper ID here

\ificcvfinal\fi
\newcommand\algoabbr{MaBi-LSTMs}
\begin{document}

%%%%%%%%% TITLE
\title{Exploring Overall Contextual Information for Image Captioning in Human-Like Cognitive Style}

\author{
Hongwei Ge$^{1}$, Zehang Yan$^1$, Kai Zhang$^1$, Mingde Zhao$^2$, Liang Sun$^1$\\
$^1$College of Computer Science and Technology, Dalian University of Technology, Dalian, China \\
$^2$Mila, McGill University, Montr\'eal, Canada \\
{\tt\small \{hwge,liangsun\}@dlut.edu.cn, mingde.zhao@mail.mcgill.ca}
%\and
}

\maketitle
\thispagestyle{empty}

%%%%%%%%% ABSTRACT
\begin{abstract}
      Image captioning is a research hotspot where encoder-decoder models combining convolutional neural network (CNN) and long short-term memory (LSTM) achieve promising results. Despite significant progress, these models generate sentences differently from human cognitive styles. Existing models often generate a complete sentence from the first word to the end, without considering the influence of the following words on the whole sentence generation. In this paper, we explore
      the utilization of a human-like cognitive style, i.e., building overall cognition for the image to be described and the sentence to be constructed, for enhancing computer image understanding. This paper first proposes a Mutual-aid network structure with Bidirectional LSTMs (MaBi-LSTMs) for acquiring overall contextual information. In the training process, the forward and backward LSTMs encode the succeeding and preceding words into their respective hidden states by simultaneously constructing the whole sentence in a complementary manner. In the captioning process, the LSTM implicitly utilizes the subsequent semantic information contained in its hidden states. In fact, MaBi-LSTMs can generate two sentences in forward and backward directions. To bridge the gap between cross-domain models and generate a sentence with higher quality, we further develop a cross-modal attention mechanism to retouch the two sentences by fusing their salient parts as well as the salient areas of the image. Experimental results on the Microsoft COCO dataset show that the proposed model improves the performance of encoder-decoder models and achieves state-of-the-art results.
\end{abstract}

%%%%%%%%% BODY TEXT
\section{Introduction}

\begin{figure}
\begin{center}
\includegraphics[width=1.0\linewidth]{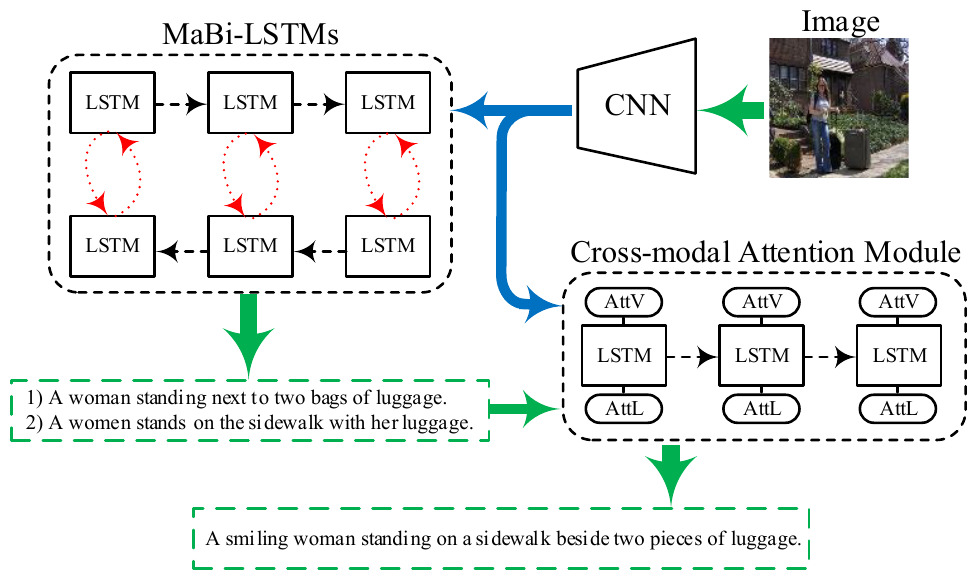}
\end{center}
   \caption{Overview of the proposed model. In the training process, image features extracted by CNN are input into \algoabbr{} to generate two sentences in an interactive manner. In the captioning process, the two sentences and image features are input into cross-modal attention module for generating the final sentence.}
\label{fig:overview}
\end{figure}

As a multimodal task, image caption generation associates with expressing image content in a sentence accurately \cite{1}, which is called `translation' from image to language \cite{2,3}. Providing an accurate description for an image is a challenging task for computers. The difficulties lie in that it not only requires the recognition of objects, attributes, and activities in an image  but also the \mbox{establishment} of fluent sentences satisfying grammatical constraints and rules.

Extensive studies have been conducted and achieve promising results by combing advanced technologies of computer vision and natural language processing. According to the different ways of generating sentences, the existing methods generally fall into three categories \cite{4}, i.e., the template-based, the transfer-based and the deep model-based methods. The template-based methods utilize multiple classifiers to recognize objects, attributes and activities in an image by mapping from image to semantics, and then fill the identified semantic words into a manually designed template to make a sentence \cite{5, 6, 7}. However, their performance noticeably deteriorates when dealing with more complex images due to the limitation of the number of classifiers and the lack of flexibility in templates. Different from the template-based methods, the transfer-based methods employ retrieving techniques to find visually similar images with the query image in an available database, and then transfer their captions to the query image \cite{8, 9, 10}. The drawback lies in that the discrepancy in the similar images leads to the inaccuracy of captioning. Recently, deep learning has increasingly attracted attention due to its practical effectiveness in difficult tasks \cite{11,12,13,14}. Thus, many deep model-based methods \cite{15,16,17,18,19} have been applied to image captioning. Generally, this kind of methods uses convolutional neural network (CNN) as a visual model to extract hierarchical features and long short-term memory (LSTM) as a language sequence model with variable length to generate descriptive sentences. Deep model-based methods not only eliminate the limitations of fixed templates but also generate original captions that are not included in available databases.

The baseline of deep model-based methods is the encoder-decoder framework where the CNN encodes raw image pixels into abstract features and the LSTM decodes abstract features into a sentence. The encoder-decoder models usually use forward LSTMs to generate words from begin to end to make a sentence \cite{37,38,39}. Recently, bidirectional LSTMs have been developed to generate sentences from two directions independently, i.e., a forward LSTM and a backward LSTM are trained  without interaction \cite{20,21}. However, there are three problems unsolved. First, only the preceding words are taken into account when predicting the next word in a sentence, namely, the information is inadequate. In fact, a sentence is made up of a sequence of words with contextual relations, where each word is related to not only its preceding words but also its subsequent words. Therefore, all explicit or implicit context information from preceding words and subsequent words will benefit the generation of current word. However, how to effectively acquire implicit information of subsequent words is challenging. Second, LSTMs may be misled during the captioning process for sentences with the same previous parts and the different remaining parts. %Second, LSTMs may be misled when inferring two sentences having the same current prefixion.
For example, considering two sentences, one is ``A little girl is happily licking an ice cream" and the other is ``A little girl is happily drinking a cup of cola".
Assuming that the current part is ``A little girl is happily", it is difficult to determine whether the next word is ``licking" or ``drinking", since the verbs cannot be determined before we get the following word to describe food or drink. Thus, when the trained LSTMs encounters this kind of situation, it would ``feel" confused when predicting ``licking" or ``drinking"  without implicit context-after information in the captioning phase. Third, due to the domain adaptability of different models, there is a definite gap between the image representation obtained by CNN and the semantic generated by LSTM. It means that the information from images cannot be utilized extensively and effectively for sentence generation \cite{32}.

To address the above challenges, we conduct research inspired by a human-like cognitive style, which builds an overall cognition for the image to be described and the sentence to be constructed. More specifically, we propose a Mutual-aid network structure with Bidirectional LSTMs (MaBi-LSTMs) for exploring overall contextual information and a cross-modal attention mechanism to bride the gap between cross-domain models. Figure~\ref{fig:overview} illustrates the captioning process by using \algoabbr{} and cross-modal attention. The main contributions of this paper lie in:

\vspace{-0.8em}
\begin{itemize}
\setlength{\itemsep}{0pt}
\setlength{\parsep}{0pt}
\setlength{\parskip}{0pt}
\item
  By designing an auxiliary structure, the forward and backward LSTMs encode the subsequent and preceding words into their respective hidden states through simultaneously constructing the whole sentence in a complementary manner. In this way, the hidden nodes contain not only the context-before information but also the context-after information.
\item
  A cross-modal attention mechanism is proposed to dynamically focus on the salient areas of images and the salient parts of the sentences. This mechanism retouches the two sentences generated by \algoabbr{} into a  higher quality sentence.
\item
  The sufficient contextual information from \algoabbr{} and the cross-modal attention mechanism alleviate the problem that LSTM cannot seamlessly utilize image features.
 \end{itemize}

\section{Related works}

As the proposed \algoabbr{} is closely related to the works utilizing deep neural networks, in this section, we review the deep models based on whether they are end-to-end trainable or phased trainable.

\subsection{End-to-end trainable models}

The end-to-end trainable models directly use images and corresponding captions for training. Vinyals et al. \cite{3} use a deep CNN to encode images and use an LSTM to decode the encoded image representation into sentences. Further, in \cite{4}, semantic features of images are taken as part of the input of LSTM to produce more accurate sentences for an image. Considering that human visual attention mechanism can filter image noise based on saliency, Xu et al. \cite{16} propose an attentive encoder-decoder model to dynamically concentrate on the local features of different image regions at each step of word sequence generation. Chen et al. \cite{25} further extend spatial attention to spatial-channel attention considering that CNN features are naturally spatial, channel-wise and multi-layer. However, the lower and fixed resolution of CNN feature maps may lead to the loss of some important local visual information. To address this problem, Fu et al. \cite{26} use selective search \cite{27} to predict image regions containing objects and feed them to CNN to extract local features. Attention mechanism that only concentrates on local features cannot model global information effectively. Yant et al. \cite{17} enhance the modelling ability of global information by a review network that plugs an LSTM in the attentive encoder-decoder. The plugged LSTM encodes local features as a series of hidden states that keep both local and global information. Although the attention mechanism can dynamically focus on salient image features, the prediction of some non-visual words (such as `to', `the', and `a') does not require any visual information but requires contextual information only. Therefore, Lu et al. \cite{28} propose an adaptive attention mechanism with visual sentinel. The visual sentinel is responsible for eliciting context information, while the CNN is responsible for extracting visual information. In order to fully exploit the long-term dependency of LSTM, Chen et al. \cite{19} regularize LSTM by forcing its current hidden state to reconstruct the previous. This strategy enables LSTM's hidden nodes to provide more information about the past. Besides, the traditional LSTM encodes its hidden state as a vector that does not contain spatial structure. For utilizing spatial structure, Dai et al. \cite{30} adopt a two-dimensional map instead of a one-dimensional vector to represent the hidden state and validate that spatial structure can boost image caption generation.

\subsection{Phased trainable models}

Different from the end-to-end trainable models, the phased trainable models look for intermediary semantic attributes (usually the high-frequency words in the vocabulary) to bridge an image and its corresponding caption. This kind of models usually consists of two main components and is trained in several stages. The first component is used to produce most related word, and the second component is used to generate a sentence by utilizing the produced words as inputs. Many strategies have been proposed along this paradigm. After producing words by employing a set of attribute detectors based on CNN, You et al. \cite{31} introduce semantic attention to focus on salient items.

Similarly, Wu et al. \cite{32} establish a mapping to caption an image with its associate words by training a CNN in a multi-label classification senary. Fang et al. \cite{33} adopt Multiple Instance Learning (MIL) as an alternative to yield associate words for the image to be captioned. Yao et al. \cite{34} deliver the words from MIL to an LSTM in different ways and empirically claim that it is the optimal strategy for putting attributes into LSTM at the beginning. Instead of putting attributes into LSTM directly, Gan et al. \cite{35} develop a semantic compositional network to exploit attributes more efficiently. The network works by extending each parameter matrix of LSTM into a set of matrices, with each being weighed based on different attributes. Besides, some other eligible features are also employed as intermediary attributes for the phased trainable models. For example, in \cite{18}, the object-level features are extracted by a pre-trained faster R-CNN \cite{36} as intermediary attributes.

Despite significant progress, the way these models generating sentences for image captioning is different from that of human beings. When people describe pictures, they first have an overall cognition for the images to be described and the sentences to be constructed. Most existing models generate a word sequence one by one in a front-to-back manner, without considering the influence of the subsequent words on the whole sentence generation. Bidirectional LSTMs have been developed to generate sentences from two directions \cite{20,21} independently. Essentially, it is the same way as before since the forward and backward LSTMs are still trained without interaction. This paper proposes a mutual-aid network structure with bidirectional LSTMs for exploring overall contextual information. In \algoabbr{}, the forward and backward LSTMs encode the succeeding and preceding words into their respective hidden states by constructing the whole sentence in a complementary manner simultaneously. The \algoabbr{} generate two sentences in an interactive way. Moreover, a special module of cross-modal attention mechanism is designed to retouch the two sentences into a new sentence with higher quality. The \algoabbr{} work in an end-to-end trainable manner, while the stage of sentence retouching works in a phased trainable manner, where the two pre-generated sentences from \algoabbr{} operate as the intermediary attributes.

\section{Methods}

In this section, we start by briefly describing the generic encoder-decoder image captioning framework, and then we give the details of the proposed model focusing on the mutual-aid network structure with bidirectional LSTMs and the cross-modal attention for decoding.

\begin{figure}
\begin{center}
\includegraphics[width=1.0\linewidth]{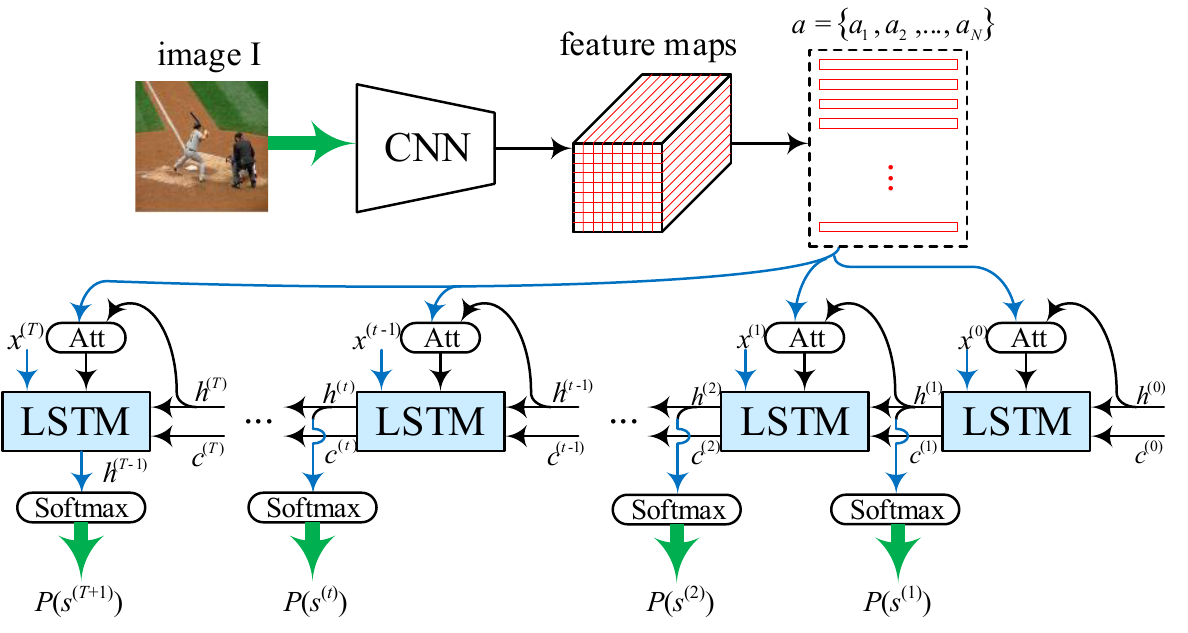}
\end{center}
   \caption{Attentive encoder-decoder diagram. Encoding stage:  the image $I$ is input into a CNN to extract the feature maps, and then the feature maps are split into a set of image features along spatial dimensions. Decoding stage: LSTM receives the word embedding vector of the previous moment $x^{(t-1)}$  and the salient image features $z^{(t)}$ from the attention module. Then, the hidden state of LSTM $h^{(t)}$ is input into the softmax layer to produce the probability distribution of words at the current moment $P(s^{(t)})$.}
\label{fig:attlstm}
\end{figure}

\subsection{Attentive encoder-decoder model}

The attention-based encoder-decoder model consists of a CNN-based encoder and a LSTM-based decoder with the attention module.

{\bf Encoder.} After feeding an image $I$ to a CNN, the feature map from the last convolutional layer is split along spatial dimensions and yields a set of vectors $a=\{a_{1}, a_{2}, ..., a_{i}, ..., a_{N}\}$ , where $a_{i}$ is a D-dimensional vector representing the feature of an image region. The process means that the CNN encodes image $I$ into $N$ vectors:
\begin{equation}
a=CNN(I)
\end{equation}

{\bf Decoder.} The word sequence $S=(s^{(0)}, s^{(1)}, ..., s^{(t)}, ...\\, s^{(T+1)})$  represents the sentence corresponding to image $I$, where $s^{(t)}$  is a one-hot vector whose length is equal to the size of the dictionary. $s^{(t)}$  indicates the index of the word in the dictionary.  $s^{(t)}$ can be embedded in a compact vector space by multiplying an embedding matrix $E$, i.e., $x^{(t)}=Es^{(t)}$. During the process of decoding $a=\{a_{1}, a_{2}, ..., a_{i}, ..., a_{N}\}$  into $S=(s^{(0)}, s^{(1)}, ..., s^{(t)}, ..., s^{(T+1)})$  , the input of LSTM at time $t$  includes not only $x^{(t-1)}$  but also the salient image feature $z^{(t)}$ computed by performing attention mechanism on all the image features:
\begin{equation}
h^{(t)}=LSTM(x^{(t-1)}, z^{(t)}, h^{(t-1)})
\end{equation}

{\bf Attention.} Visual attention is usually implemented by taking the weighted sums of all image features to obtain the salient image features, which is denoted as $z^{(t)}=Att(h^{(t-1)}, a)$, where $z^{(t)}$ is correlated with  $h^{(t-1)}$. The detailed computation is formulized as follows:
\begin{equation}
z^{(t)}=\sum_{i=1}^{N}\alpha_{i}^{(t)}a_{i}
\end{equation}
\vspace{-0.5em}
\begin{equation}
\alpha_{i}^{(t)}=\frac{exp(MLP(h^{(t-1)}, a_{i}))}{\sum_{j=1}^{N}exp(MLP(h^{(t-1)}, a_{j}))}
\end{equation}
Where $\alpha_{i}^{(t)}$  represents the weight of the feature of the $i$-th image region at time $t$, and $MLP$ represents a multi-layer perceptron.

A softmax layer is employed to convert $h^{(t)}$  into the probability distribution of $s^{(t)}$, from which the word at time $t$ can be sampled:
\begin{equation}
P(s^{{(t)}}|s^{{(0)}}, ..., s^{{(t-1)}}, I, \theta)=softmax(h^{(t)})
\end{equation}
where $\theta$ denotes the learnable parameters of LSTM.

Figure~\ref{fig:attlstm} shows the structure of the attentive encoder-decoder model and details the interaction between the attention module and LSTM. To illustrate more clearly, the subsequent diagrams will use an abbreviated form $LSTMA$ to represent LSTM with the attention module. Thus, \mbox{Equation (2)} is rewritten as:
\begin{equation}
\begin{aligned}
h^{(t)} &= LSTM(x^{(t-1)}, Att(h^{(t-1)}, a), h^{(t-1)}) \\
&= LSTMA(x^{(t-1)}, a, h^{(t-1)})
\end{aligned}
\end{equation}

\subsection{Mutual-aid network structure with bidirectional LSTMs}

\begin{figure}
\begin{center}
\includegraphics[width=1.0\linewidth]{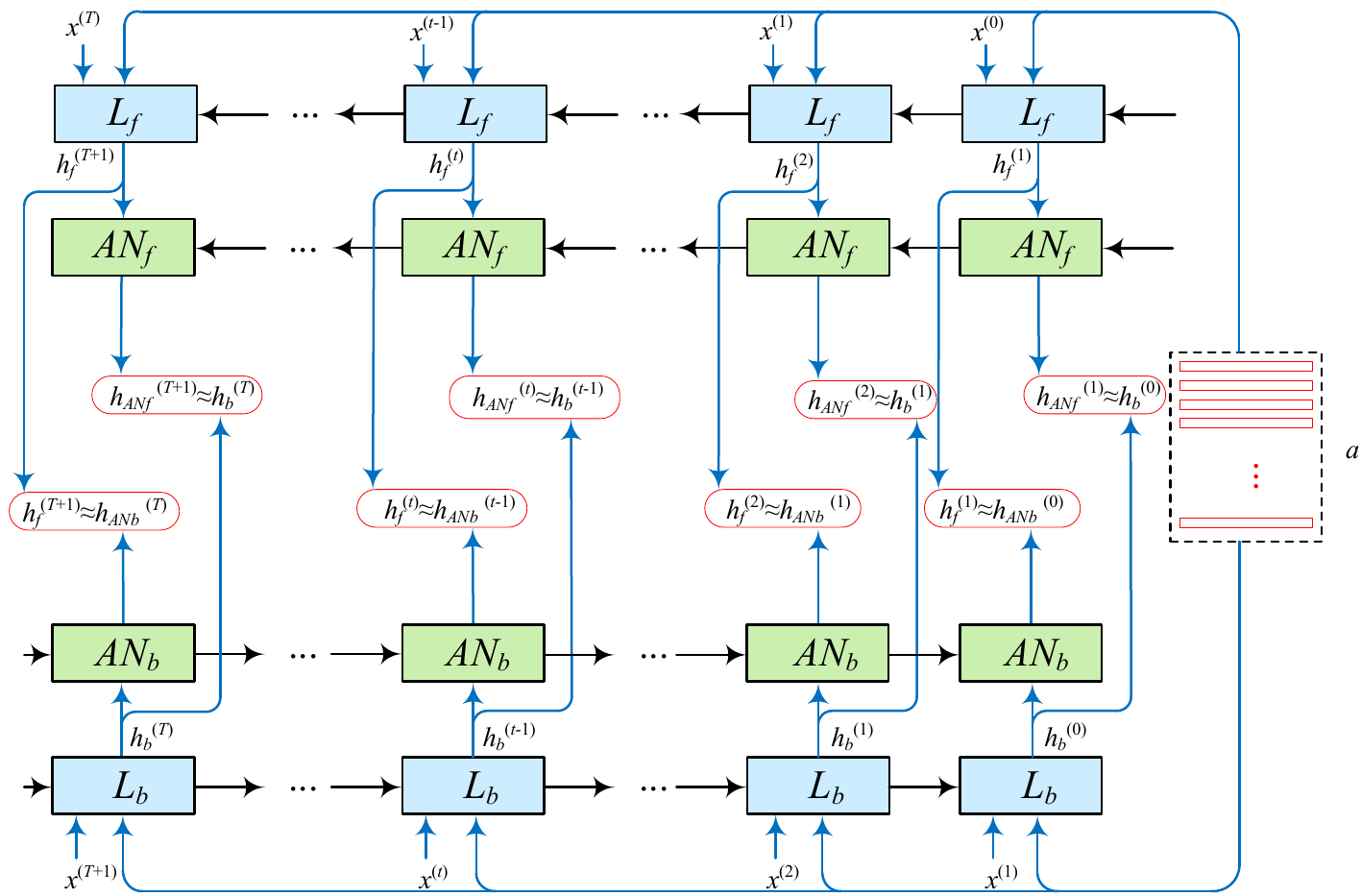}
\end{center}
   \caption{The schematic diagram of the proposed \algoabbr{}. The green boxes denote the auxiliary modules, and the blue boxes denote the original bidirectional LSTMs. $L_{f}$  and $L_{b}$  reserve two sequences of hidden states $h_{f}=(h_{f}^{(1)}, ..., h_{f}^{(t)}, ..., h_{f}^{(T+1)})$  and  $h_{b}=(h_{b}^{(0)}, ..., h_{b}^{(t-1)}, ..., h_{b}^{(T)})$ in the process of generating sentences by using the image feature set $a$. In addition, $L_{f}$ and $L_{b}$ construct the hidden states of each other. $AN_{f}$  models $h_{f}$  in an forward order to build the new state sequence $h_{AN_{f}}$ to approximate $h_{b}$ for capturing the context-after information. $AN_{b}$  traces $h_{b}$  in a backward order to form state sequence $h_{AN_{b}}$  to approximate $h_{f}$ for capturing the context-before information.}
\label{fig:malstm}
\end{figure}

In this section, we present the details of the proposed \algoabbr{}. The schematic diagram of \algoabbr{} is shown in Figure~\ref{fig:malstm}.

We introduce the auxiliary forward aid network $AN_f$ and backward aid network $AN_b$ in the bidirectional LSTMs. Then the auxiliary and the original LSTMs (foward LSTM $L_{f}$ and backward LSTM $L_{b}$) work coordinately by mutually constructing hidden states via the auxiliary LSTMs. By introducing the auxiliary LSTMs (forward aid network $AN_{f}$ and backward aid network $AN_{b}$) structure into the bidirectional LSTMs, the auxiliary and the original LSTMs (foward LSTM $L_{f}$ and backward LSTM $L_{b}$) work coordinately by mutually constructing hidden states.

In \algoabbr{}, $L_{f}$  constructs the hidden states of $L_{b}$ via $AN_{f}$ to capture latent context-after information and $L_{b}$  builds  the hidden states of $L_{f}$ via $AN_{b}$ to capture latent context-before information. To effectively construct the hidden states of $L_{f}$ and $L_{b}$, we adopt LSTM as the specific implementation for $AN_{f}$ and $AN_{b}$ to model the temporal dynamics of time series data. In the process of sentence generation, $L_{f}$  updates its hidden states in a front-to-back manner:
\begin{equation}
\begin{aligned}
h_{f}^{(t)}\!=\!L_{f}(x^{(t-1)}\!,\! a\!,\! h_{f}^{(t-1)})
\!=\!LSTMA_{f}(x^{(t-1)}\!,\! a\!,\! h_{f}^{(t-1)})
\end{aligned}
\end{equation}
$L_{b}$  updates its hidden states from back to front:
\begin{equation}
\begin{aligned}
h_{b}^{(t-1)}=L_{b}(x^{(t)}, a, h_{b}^{(t)})
=LSTMA_{b}(x^{(t)}, a, h_{b}^{(t)}))
\end{aligned}
\end{equation}
To make $L_{f}$  and $L_{b}$  aid each other,  $AN_{f}$ takes  $h_{f}^{(t)}$ as input to generate hidden states  $h_{AN_{f}}^{(t)}$ for approximating $h_{b}^{(t-1)}$:
\begin{equation}
\begin{aligned}
h_{AN_{f}}^{(t)}
= AN_{f}(h_{f}^{(t)}, h_{AN_{f}}^{(t-1)}) \approx h_{b}^{(t-1)}
\end{aligned}
\end{equation}
Similarly, $AN_{b}$ uses $h_{b}^{(t-1)}$ to approximate $h_{f}^{(t)}$:
\begin{equation}
\begin{aligned}
h_{AN_{b}}^{(t-1)}
= AN_{b}(h_{b}^{(t-1)}, h_{AN_{b}}^{(t)}) \approx h_{f}^{(t)}
\end{aligned}
\end{equation}

In Figure~\ref{fig:malstm},  $L_{f}$ and $L_{b}$  construct the hidden states of each other via $AN_{f}$  and $AN_{b}$,  respectively. The  $h_{AN_{b}}^{(t-1)}$ integrates the hidden states of $L_{b}$ after time $t-1$ (including $h_{b}^{(t-1)}$  up to $h_{b}^{T}$ ). The $h_{f}^{(t)}$  is closely related to the first half of the sentence (from $x^{(0)}$  to $x^{(t-1)}$ ), which contains the context-before information implicitly. If we can make $h_{AN_{b}}^{(t-1)}$  close enough to $h_{f}^{(t)}$ ,  $h_{AN_{b}}^{(t-1)}$ will contain extra knowledge about the first half sentence, which is transferred to $h_{b}^{(t-1)}$ implicitly. Therefore, it helps to predict the next word $x^{(t-1)}$  based on  $h_{b}^{(t-1)}$. By analogy, $h_{AN_{f}}^{(t)}$  integrates  the hidden states of $L_{f}$ before time $t$ (including $h_{f}^{(1)}$ to $h_{f}^{(t)}$ ). The  $h_{b}^{(t-1)}$ is closely related to the latter part of the sentence (from $x^{(t)}$  to $x^{(T+1)}$), which contains context-after information implicitly . If we can make  $h_{AN_{f}}^{(t)}$ close enough to $h_{b}^{(t-1)}$ , $h_{AN_{f}}^{(t)}$  will contain extra knowledge about the latter part of the sentence. Hence, it makes sense to predict the next word $x^{(t)}$  based on $h_{f}^{(t)}$  with higher confidence.

{\bf Loss function.} $L_{f}$  and $L_{b}$  offer two probability distributions bidirectionally. Therefore, the first part of the loss function should be the sum of two negative log-likelihood functions that represents the errors via supervised training. To make the $AN_{f}$  and the $AN_{b}$ effectively model the hidden states of $L_{f}$  and  $L_{b}$, respectively, the least square error that represents the error of construction is used as the second part of the loss function. The final loss function is the sum of $L_{1}$ and $L_{2}$ weighted by $\lambda$.

\begin{figure}
\begin{center}
\includegraphics[width=1.0\linewidth]{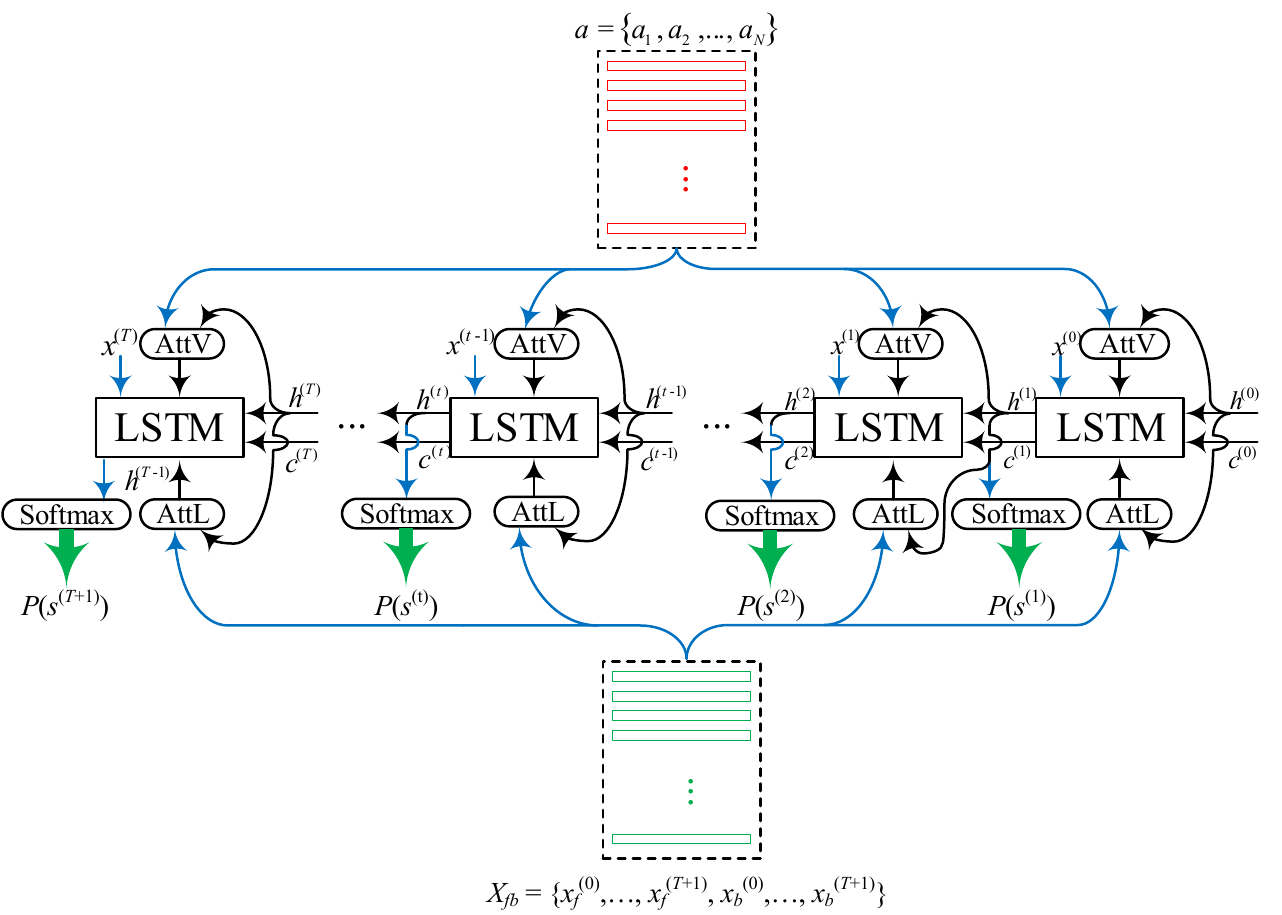}
\end{center}
   \caption{The cross-modal attentive decoder. Cross-modal attention incorporates semantic attention AttL and visual attention AttV. The cross-modal attention module selects salient image features and semantic features for updating the hidden states followed by the softmax to generate word probability distributions.}
\label{fig:mattlstm}
\end{figure}

\begin{equation}
\begin{aligned}
L_{1}=&-\sum_{t=1}^{T}log(P_{f}(s^{(t)}|s^{(0)}, ...,s^{(t-1)}, I, \theta_{f}))\\
&-\sum_{t=1}^{T}log(P_{b}(s^{(t)}|s^{(T+1)}, ...,s^{(t+1)}, I, \theta_{b}))
\end{aligned}
\end{equation}
\begin{equation}
L_{2}=\sum_{t=1}^{T}\Arrowvert h_{AN_{b}}^{(t-1)} - h_{f}^{(t)} \Arrowvert^{2} + \sum_{t=1}^{T}\Arrowvert h_{AN_{f}}^{(t)} - h_{b}^{(t-1)} \Arrowvert^{2}
\end{equation}
\begin{equation}
L_{total}=L_{1}+\lambda L_{2}
\end{equation}

\subsection{Cross-modal attention for decoding}

Cross-modal attention extends visual attention to both visual and semantic aspects. From the visual aspect, CNN can extract image feature set $a$. From the semantic aspect, the trained $L_{f}$  and $L_{b}$  can produce two sentences  $S_{f}=(s_{f}^{(0)}, ..., s_{f}^{(T+1)})$ and $S_{b}=(s_{b}^{(0)}, ..., s_{b}^{(T+1)})$ . $S_f$ and $S_b$ can be represented by their corresponding word embedding $X_{fb}=(x_{f}^{(0)}, ..., x_{f}^{(T+1)}, x_{b}^{(0)}, ..., x_{b}^{(T+1)})$ . The image feature set $a$  and the word embedding $X_{fb}$  constitute the multimodal features in the cross-modal environment. Figure~\ref{fig:mattlstm} shows the process of generating the final sentence by using cross-modal attention.

Cross-modal attention can be disassembled into visual attention and semantic attention and can operate on image features and word embeddings to improve the saliency of cross-modal features. The salient image features and the word embedding obtained by cross-modal attention are concatenated, and then input into LSTM to update the hidden states. Through the softmax layer, the hidden states at each time step are converted into probability distributions of words, where the whole sentence can be predicted by sampling. The loss function here is taken as the negative log-likelihood.

\begin{table}
\begin{center}
\resizebox{1\hsize}{!}{
\begin{tabular}{l|c|c@{ }c@{ }c@{ }c@{ }c@{ }c@{ }c}
\toprule
\toprule
LSTM & $\lambda$ & BLEU-1 & BLEU-2 & BLEU-3 & BLEU-4  & METEOR  & ROUGE-L  & CIDEr \\
\hline
\multirow{6}*{$L_{f}$} & 0 & 72.7 & 56.0 & 42.4 & 32.1 & 25.5 & 53.6 & 99.9 \\
~ & 0.001 & 73.8 & 57.4 & 43.7 & 33.1 & 25.7 & 54.5 & 102.6 \\
~ & 0.005 & 73.9 & 57.6 & 43.9 & 33.2 & 25.8 & \bf{54.9} & 102.6 \\
~ & 0.01 & \bf{74.4} & \bf{58.0} & \bf{44.2} & \bf{33.5} & \bf{26.1} & \bf{54.9} & \bf{105.7} \\
~ & 0.05 & 73.8 & 57.2 & 43.7 & 33.4 & 25.7 & 54.3 & 102.6 \\
~ & 0.1 & 72.8 & 56.1 & 42.4 & 32.0 & 25.3 & 53.8 & 100.0 \\
\hline
\multirow{6}*{$L_{b}$} & 0 & 72.9 & 56.0 & 41.8  & 31.0 & 25.3 & 53.5 & 99.7 \\
~ & 0.001 & 73.5 & \bf{57.3} & 43.0  & \bf{32.2} & \bf{25.6} & 54.2 & 102.5 \\
~ & 0.005 & 73.2 & 56.7 & 42.7  & 31.8 & 25.5 & \bf{54.3} & 102.3 \\
~ & 0.01 & \bf{74.2} & \bf{57.3} & \bf{43.9}  & 31.9 & 25.5 & 54.1 & \bf{103.6} \\
~ & 0.05 & 73.2 & 56.7 & 42.8  & 31.9 & 25.4 & 53.9 & 102.2 \\
~ & 0.1 & 72.8 & 55.7 & 41.5  & 30.7 & 25.2 & 53.3 & 99.5 \\
\toprule
\toprule
\end{tabular}
}
\end{center}
\caption{Results of the attentive encoder-decoder models with different weight parameters. When $\lambda = 0$, the model is without the auxiliary structure.}
\label{tab:lambda}
\end{table}

\section{Experiments}

To validate the effectiveness of the proposed algorithm, we conduct experiments on the Microsoft COCO dataset \cite{40}. The experimental results are analysed and compared with the state-of-the-art algorithms.

\subsection{Dataset and evaluation metrics}

The Microsoft COCO dataset is a popular large scale dataset for image captioning, including 82,783 images for training, 40,504 images for validation, and 40,775 images for testing. Each image in the training and validation set accompanies with five sentences and each sentence can describe the image contents. Such descriptive sentences are manually labelled by humans through the Amazon Mechanical Turk platform. To make fair comparisons with other methods, we follow a common accepted configurations in the community \cite{1}, and select 5000 images from the validation set for validating and another 5,000 images from the validation set for testing.

Commonly used metrics for measuring caption quality include BLEU-n \cite{43}, ROUGE-L \cite{45}, METEOR \cite{44}, and CIDEr \cite{46}. BLEU-n measures the similarity between a candidate sentence against reference sentences by computing the n-gram precision. METEOR computes not only uni-gram precision but also recall and uses their weighted harmonic mean. ROUGE-L employs the longest common subsentences to measure the similarity between a candidate sentence and reference sentences at the sentence level. The above three evaluation metrics are derived from the study of machine translation tasks, while CIDEr is based on human consensus and specifically designed for image captioning. To validate the effectiveness of the proposed method, we use the four evaluation metrics: BLEU-n (n=1, 2, 3, 4), METEOR, ROUGE-L and CIDEr.

\subsection{Implementation details}

After using the Stanford PTB Tokenizer \cite{42} to split sentences into words, we remove all the punctuations. The dictionary is then built by words with the frequency higher than 5. Besides those high-frequency words, the dictionary contains three particular tokens: the sentence-starting token $<start>$, the sentence-ending token $<end>$, and the unknown token $<unk>$, where $<unk>$ is used to replace words with the frequency lower than 5. For extracting compressed image features and reducing computational cost, features from CNN are input into a one-layer neural network with linear activation function for dimension reduction. To alleviate over-fitting, we stop training when the CIDEr score on the validation set begins to decline. There exists different strategies when continuously sampling words from probability distributions at each time step. The most straightforward approach is to pick the word with the highest probability at each time step to make a sentence. However, such a strategy is greedy. Beam Search differs from Greedy Search in selecting $n$ candidate words with the highest probabilities. Beam Search degenerates into Greedy Search when $n = 1$ and transforms into Breadth-First Search when $n$ takes the maximum value. Considering both the computation and the performance, we set the value of $n$ as $3$.

The proposed algorithm is implemented by Python and Theano on a workstation with a Tesla P40 GPU.

\begin{table}
\begin{center}
\resizebox{1\hsize}{!}{
\begin{tabular}{@{ }l|c@{ }c@{ }c@{ }c@{ }c@{ }c@{ }c@{ }}
\toprule
\toprule
Methods & BLEU-1 & BLEU-2 & BLEU-3 & BLEU-4  & METEOR  & ROUGE-L  & CIDEr \\
\hline
DeepVS \cite{1} & 62.5 & 45.0 & 32.1 & 23.0 & 19.5 & - & 66.0 \\
MSR \cite{33} & - & - & - & 25.7 & 23.6 & - & - \\
gLSTM \cite{4} & 67.0 & 49.1 & 35.8 & 26.4 & 22.7 & - & 81.3 \\

Bi-S-LSTM$^V$ \cite{21}     &68.7    &50.9   &36.4   &25.8   &22.9   &-     &73.9       \\
Bi-LSTM$^{A,+M}$ \cite{21}  &65.6    &47.4   &33.3   &23.0   &21.1   &-     &69.5       \\

Attr-CNN+LSTM \cite{32} & 74.0 & 56.0 & 42.0 & 31.0 & 26.0 & - & 94.0 \\
ATT-FCN \cite{31} & 70.9 & 53.7 & 40.2 & 30.4 & 24.3 & - & - \\
Soft-Attention \cite{16} & 70.7 & 49.2 & 34.4  & 24.3 & 23.9 & - & - \\
Hard-Attention \cite{16} & 71.8 & 50.4 & 35.7  & 25.0 & 23.0 & - & - \\
Review Net \cite{17} & - & - & -  & 29.0 & 23.7 & - & 88.6 \\
LSTM-A5 \cite{34} & 73.0 & 56.5 & 42.9  & 32.5 & 25.1 & 53.8 & 98.6 \\
SCA-CNN \cite{25} & 71.9 & 54.8 & 41.1  & 31.1 & 25.0 & 53.1 & 95.2 \\
GBVS \cite{51} &-&-&-&28.7&23.5&21.2&84.1\\
Areas of Attention \cite{50}&-&-&-&31.9&25.2&-&98.1\\
ARNet \cite{19} & 74.0 & 57.6 & 44.0  & 33.5 & 26.1 & 54.6 & 103.4 \\
SCN-LSTM \cite{35} & 72.8 & 56.6 & 43.3  & 33.0 & 25.7 & - & 101.2 \\
Skeleton \cite{48} & 74.2 & 57.7 & 44.2  & 34.0 & 26.8 & 55.2 & 106.9 \\
AdaAtt \cite{28} & 74.2 & 58.0 & 43.9  & 33.2 & 26.6 & 54.9 & 108.5 \\
GroupCap \cite{39}          &74.4   &58.1   &44.3   &33.8   &26.2   &-      &-          \\
NBT \cite{49} & 75.5 & - & -  & 34.7 & 27.1 & - & 107.2 \\
\hline
Our MaBi-LSTMs($L_{f}$) &  77.5 & 59.9 & 46.2  & 36.4 & 27.2 & 56.1 & 110.4 \\
Our MaBi-LSTMs($L_{b}$) & 75.8 & 59.4 & 45.1  & 34.9 & 26.8 & 55.8 & 108.5 \\
Our MaBi-LSTMs(with attention) & \bf{79.3} & \bf{61.2} & \bf{47.3}  & \bf{36.8} & \bf{28.1} & \bf{56.9} & \bf{116.6} \\
\toprule
\toprule
\end{tabular}
}
\end{center}
\caption{Comparative results of MaBi-LSTMs with different algorithms on the COCO test set.}
\label{tab:finetune}
\end{table}

\subsection{Experiments for \algoabbr{}}

To verify the effectiveness of mutual-aid bidirectional LSTMs, we conduct the comparative experiments of the attentive encoder-decoder models with or without the auxiliary structure. An Inception-V4 \cite{41} pre-trained on the ImageNet classification dataset acts as the encoder and the parameters are kept unchanged. The loss function of the mutual-aid network structure with bidirectional LSTMs consists of two items: the cross-entropy error $L_{1}$ and the least squares error $L_{2}$. The parameter $\lambda$ is important for adjusting the ratio of these two items. We empirically find that the value of $L_{2}$ is 2 or 3 orders higher than that of $L_{1}$. In order to balance these two items so that they can work on the same level, we search for an appropriate value for $\lambda$ in the interval of $[0, 0.1]$. Actually, we test the performances of the proposed model when the values of $\lambda$  is taken as 0, 0.001, 0.005, 0.01, 0.05, and 0.1 respectively, and the results are presented in Table~\ref{tab:lambda}.

When $\lambda$  is taken as 0, the least squares error does not work, and only the cross-entropy error is used, that is, without auxiliary structure, the training of $L_{f}$  and $L_{b}$ has no interaction. When  $\lambda$ is taken as non-zero values except 0.1, the scores are obviously higher than those with $\lambda = 0$. The results verify the effectiveness of the auxiliary structure. When $\lambda$ is taken as $0.1$, the least squares error exceeds cross entropy excessively according to the difference in magnitude, which makes the training target of the model more biased towards the mutual construction and ignores the ability of word prediction. In addition, $L_{f}$ generally performs better than $L_{b}$ for each metric. This is consistent with our common sense because it is easier to recite a sentence from front to back than to recite it reversely. When $\lambda$ is taken as $0.01$, all the metrics generally achieve the best values. So, we set $\lambda$ as $0.01$ in the remaining experiments.

The results listed in Table~\ref{tab:lambda} are obtained by only training the decoder while keeping the encoder frozen. To further improve the model performance, the model is continued to be trained  by jointly fine-tuning the encoder and decoder. Table~\ref{tab:finetune} presents the results obtained by $L_{f}$ and $L_{b}$ and the comparison results with some state-of-the-art algorithms. Notably, we determine whether to stop the training process in time for alleviating over-fitting based on CIDEr score. Without more complicated training tricks and experiences, the proposed model still achieves competitive results.

\begin{figure*}
\begin{center}
\includegraphics[width=0.90\linewidth]{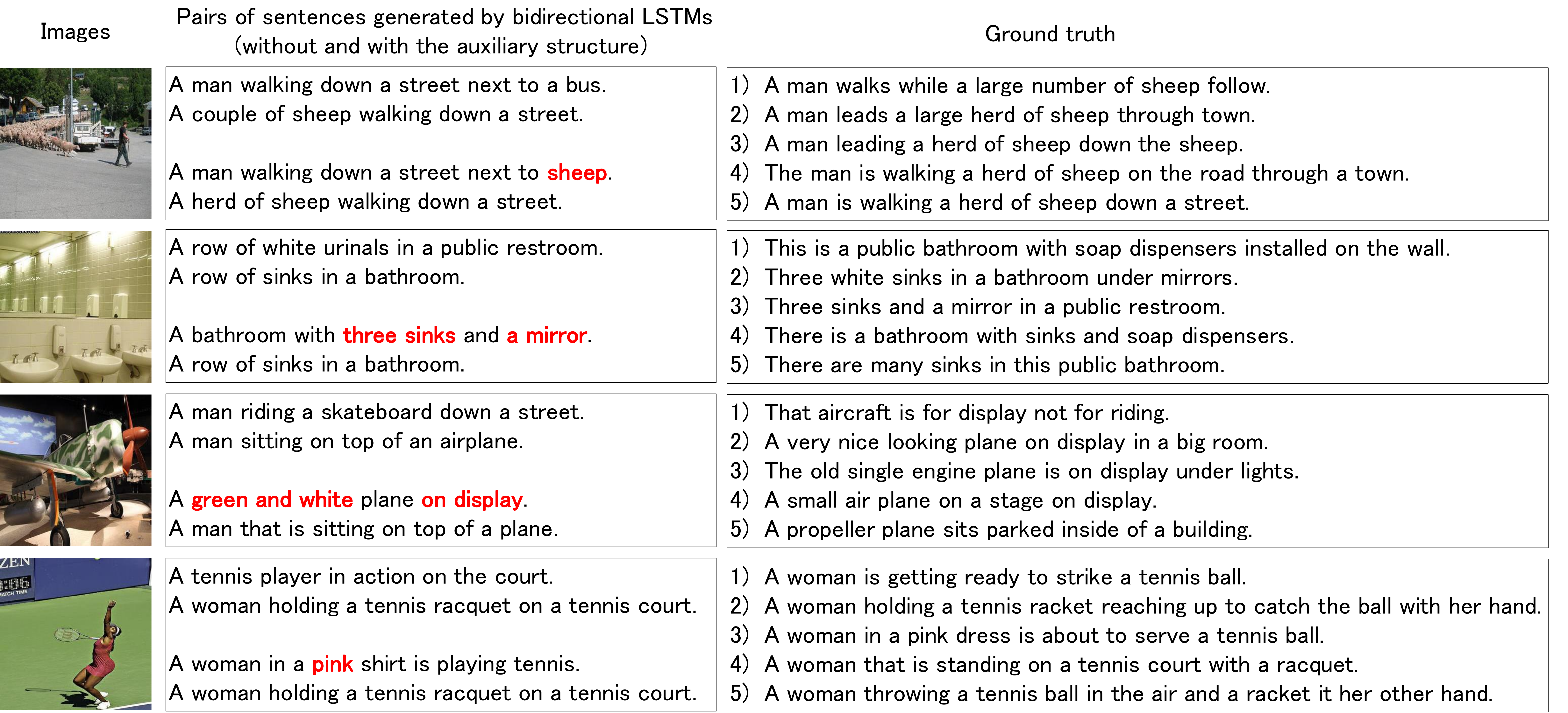}
\end{center}
   \caption{Pairs of captions generated by the proposed model with and without the auxiliary structure. It can be observed that \algoabbr{} can produce more detailed and accurate words. Those words are marked in a bold red font, such as `green and white', `pink', `three', etc.}
\label{fig:examples_mblstm}
\end{figure*}

\begin{figure*}
\begin{center}
\includegraphics[width=0.98\linewidth]{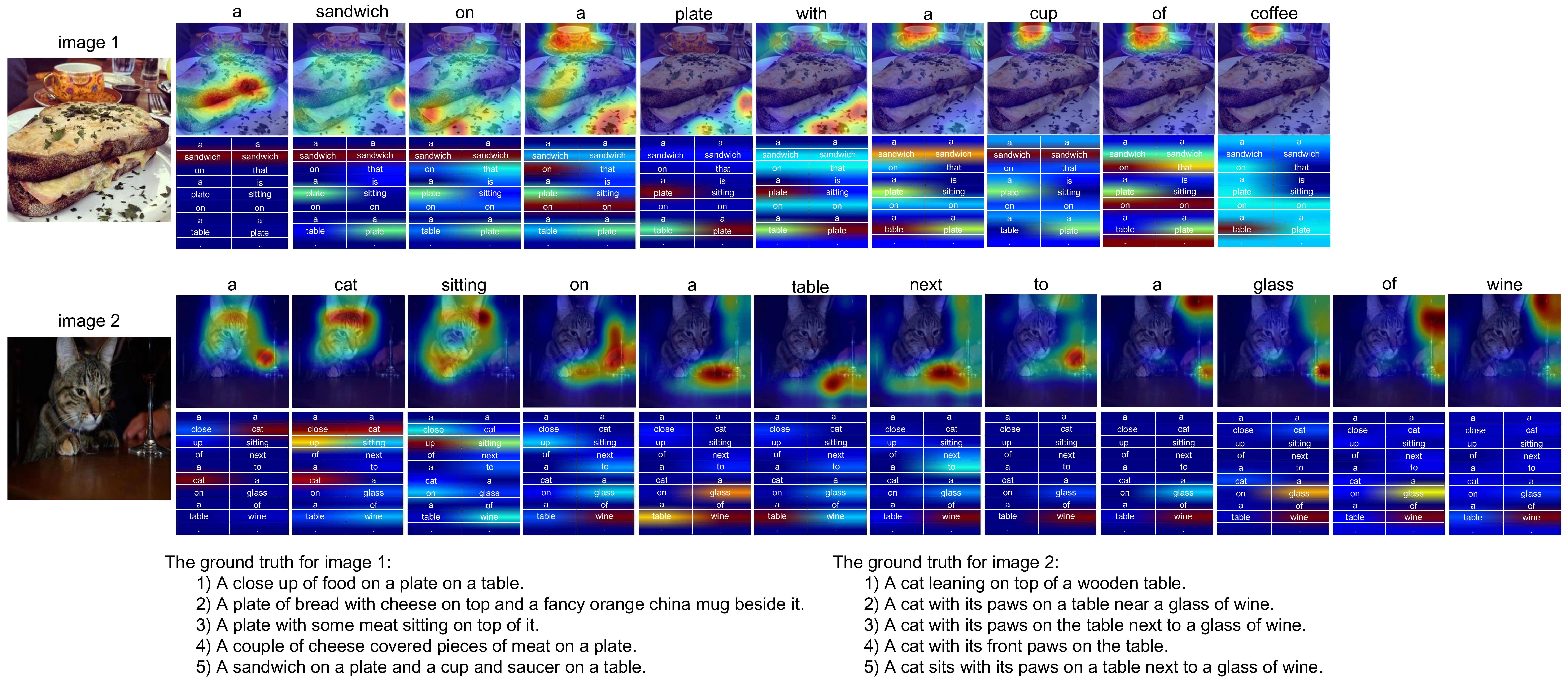}
\end{center}
   \caption{Cross-modal attention visualization. The visualization shows the dynamic changes of attention weights while generating captions. The first and second rows of each example visualize the attention weights on images and the pre-generated sentences. The left and right columns in the second row represent the two pre-generated sentences. The redder the greater weight, the bluer the smaller weight.}
\label{fig:examples_atten}
\end{figure*}

Figure~\ref{fig:examples_mblstm} shows some pairs of captions generated by $L_{f}$ and $L_{b}$ with and without the auxiliary structure for 4 images. From the comparison results, it can be seen that \algoabbr{} can improve the quality of image captions. For the first image, either of the captions generated by the model without the auxiliary structure is incomplete. One only describes the person and the other only describes sheep. \algoabbr{} can capture both objects and express the person and sheep in a single sentence. For the second image, the initial captions do not include the important image content, `mirror', and cannot accurately identify the number of sinks. The captions generated by using the auxiliary structure can describe `mirror' and the number of sinks. For the third image, it is commendable that \algoabbr{} can still produce reliable captions even if the captions generated without auxiliary structure are completely unqualified in describing the image contents. For the fourth image, although the captions describe the main objects in image contents, they can be further polished by adding an adjective word, such as `pink'.

In summary, \algoabbr{} can not only accurately describe the objects contained in the image, but also correct some of the wrong descriptions. Moreover, the information provided by some fine qualifiers, such as colours, can also be supplemented by \algoabbr{}.

\subsection{Experiments for the overall model}
\algoabbr{} generate a pair of captions for an image in both forward and backward directions. Those words in the pre-generated captions are converted into word embedding vectors. Word embeddings and the image features extracted by the CNN are input into cross-modal attention module for producing image captions with higher quality. Following the same strategy as the previous experiments, we first freeze the parameters of the CNN to train only the LSTM with cross-modal attention and then train the whole model by fine-tuning jointly. The last row in Table~\ref{tab:finetune} presents the results obtained by using \algoabbr{} and cross-modal attention together. It can be seen that cross-modal attention further enhances caption quality based on the pre-generated captions.  Among all the compared methods, the overall model with cross-modal attention yields the best results.

Figure~\ref{fig:examples_atten} illustrates two examples to visually demonstrate the effectiveness of cross-modal attention for image captioning. Because multimodal attention consists of visual attention and semantic attention, attention weights can be put on visual information, semantic information or their combination to provide more comprehensive hints for the generation of the next word.

For the first example, the two pre-generated captions do not describe `cup'. Therefore, no useful information can be obtained from semantic attention when generating `a cup of coffee'. However, our model can pay attention to the image area containing the cup through visual attention. When generating `sandwich', semantic attention focuses on the pre-generated `sandwich' despite that visual attention does not seem to work. To generate `plate', cross-modal attention focuses on the image area containing the plate and the pre-generated  `plate' simultaneously. %It indicates that cross-modal attention has a certain degree of redundant correcting function.
In the second example, one pre-generated caption misses `wine glass' and the other lacks `table'. The combination of two sentences contains the complete image contents. In this case, our model turns to focus on the related image regions and semantic words simultaneously so that the model can produce a caption containing the complete image contents.

\section{Conclusion}

Inspired from the natural expression abilities of human beings in picture descriptions, the paper proposes a mutual-aid network structure with bidirectional LSTMs for acquiring overall contextual information. By designing an auxiliary structure,  \algoabbr{} make full use of context information by mutually constructing hidden states.  To the best of our knowledge, this is the first time to explore the mutual-aid structure of bidirectional LSTMs. Moreover, this paper introduces a cross-modal attention mechanism by combining visual attention and semantic attention for bridging the gap between cross-domain models. The experimental results show that the proposed algorithm has the competitive performance for image captioning. Importantly, our interests are placed on exploring the human-like cognitive style for image captioning. The methodology can be integrated into other deep learning models and applied to machine translation as well as vision question answering.

\noindent {\bf Acknowledgments.} This work is supported by the National Key R\&D Program of China (2018YFB1600600), the National Natural Science Foundation of China (61572104), the Dalian Science and Technology Innovation Fund (2019J12GX035), and the Key Program of the National Natural Science Foundation of China (U1808206).

{\small
\bibliographystyle{ieee}
\bibliography{egbib}
}

\end{document}